\documentclass[letterpaper, 10 pt, journal, twoside]{IEEEtran}  

\usepackage{graphicx}
\usepackage{amsmath}
\usepackage{amssymb}
\usepackage{booktabs}
\usepackage{color}
\usepackage{hyperref}
\usepackage{xspace}
\usepackage{url}
\usepackage{multirow}
\usepackage{booktabs}
\usepackage{colortbl}
\usepackage{latexsym}
\usepackage{hyperref}
\usepackage{epsfig}
\usepackage{latexsym}
\usepackage{lscape}
\usepackage{amsfonts}
\usepackage{listings}
\usepackage{mathdots}
\usepackage{subfig}
\usepackage{xcolor}

\IEEEoverridecommandlockouts

\definecolor{seagreen}{rgb}{0.76,1.0,0.76}  
\definecolor{greyblue}{rgb}{0.7,0.81,1.0} 
\definecolor{lightblue}{rgb}{0,0.69,0.94} 
\definecolor{myRed}{rgb}{1.0,0,0} 

\newcommand {\tabincell} [2] {\begin{tabular}{@{}#1@{}}#2\end{tabular}}
\newcommand {\aligntablebestresult} [1] {\cellcolor{seagreen}\textbf{#1}}
\newcommand {\tablebestresult} [1] {\cellcolor{seagreen}\textbf{#1}}
\newcommand \greyblueback {\bgroup\markoverwith
  {\textcolor{greyblue}{\rule[-.5ex]{2pt}{2.5ex}}}\ULon}
\newcommand \lightblueback {\bgroup\markoverwith
  {\textcolor{lightblue}{\rule[-.5ex]{2pt}{2.5ex}}}\ULon}

\makeatletter
\DeclareRobustCommand\onedot{\futurelet\@let@token\@onedot}
\def\@onedot{\ifx\@let@token.\else.\null\fi\xspace}

\def\ie{\emph{i.e}\onedot}

\def\etal{\emph{et al}\onedot}
\makeatother

\title{$\mathcal{S}^{2}$Net: Accurate Panorama Depth Estimation on Spherical Surface}

\author{Meng Li$^{1}$, Senbo Wang$^{1}$, Weihao Yuan$^{1}$, Weichao Shen$^{1}$, Zhe Sheng$^{1}$ and Zilong Dong$^{1}$
\thanks{$^{1}$Alibaba Group, No.1008 Dengcai Street, 310000, Hangzhou City, China
{\tt\footnotesize wsb\_pro@live.com}, {\tt\footnotesize shengzhe.sz@alibaba-inc.com}, 
{\tt\footnotesize list.dzl@alibaba-inc.com}}
\thanks{Meng Li and Senbo Wang contribute equally to this paper.}
\thanks{Digital Object Identifier 10.1109/LRA.2023.3234820}
\thanks{\textcircled{c} 2023 IEEE. Personal use of this material is permitted. 
Permission from IEEE must be obtained for all other uses, in any current or future media, 
including reprinting/republishing this material for advertising or promotional purposes, 
creating new collective works, for resale or redistribution to servers or lists, 
or reuse of any copyrighted component of this work in other works.}
}

\begin{document}

\maketitle

\begin{abstract}
Monocular depth estimation is an ambiguous problem, 
thus global structural cues play an important role in current data-driven single-view depth estimation methods.
Panorama images capture the complete spatial information of their surroundings utilizing the equirectangular projection which introduces large distortion. 
This requires the depth estimation method to be able to handle the distortion and extract global context information from the image. 
In this paper, we propose an end-to-end deep network for monocular panorama depth estimation on a unit spherical surface. 
Specifically, we project the feature maps extracted from equirectangular images onto unit spherical surface sampled by uniformly distributed grids, 
where the decoder network can aggregate the information from the distortion-reduced feature maps.
Meanwhile, we propose a global cross-attention-based fusion module to fuse the feature maps from skip connection and enhance the ability to  obtain global context. 
Experiments are conducted on five panorama depth estimation datasets, 
and the results demonstrate that the proposed method substantially outperforms previous state-of-the-art methods. 
All related codes will be open-sourced in the upcoming days. 


\end{abstract}

\section{Introduction}

\IEEEPARstart{P}{redicting} the depth from only a single image is particularly challenging, as this problem is inherently ambiguous~\cite{persson2021parameterization}. 
Current data-driven methods predict depth by learning implicit structural and semantic information that provides enough monocular cues for depth estimation.
This requires the network to look at bigger parts of the scene~\cite{mertan2022single}, 
which has been addressed by many works~\cite{huynh2020guiding,bhat2021adabins} in the field of single-view depth estimation in recent years.

Previous work~\cite{huynh2020guiding} tackles the problem of extending receptive field by adding a non-local depth-attention module between encoder and decoder, but training this module requires additional supervision.
Adabins~\cite{bhat2021adabins} uses a post-processing module composed by vision transformer to refine the output of traditional encoder-decoder architecture, but the receptive field of ``traditional" parts of the network has not been enlarged.
Ranftl\etal~\cite{ranftl2021vision} introduce vision transformer as its feature extraction module, 
which gives feature maps global receptive field at every stage, resulting in substantial improvements compared with fully-convolutional network~\cite{ranftl2021vision}. 
However, the ability of encoding non-local geometric dependencies is still limited as it uses a traditional convolutional network as a decoder. 
Panorama depth estimation method SliceNet~\cite{pintore2021slicenet} uses an RNN network to enhance its capability of retrieving global geometric information, but it requires additional constraints that panorama images are gravity aligned.

\begin{figure}
    \centering 

    \includegraphics[width=0.99\linewidth]{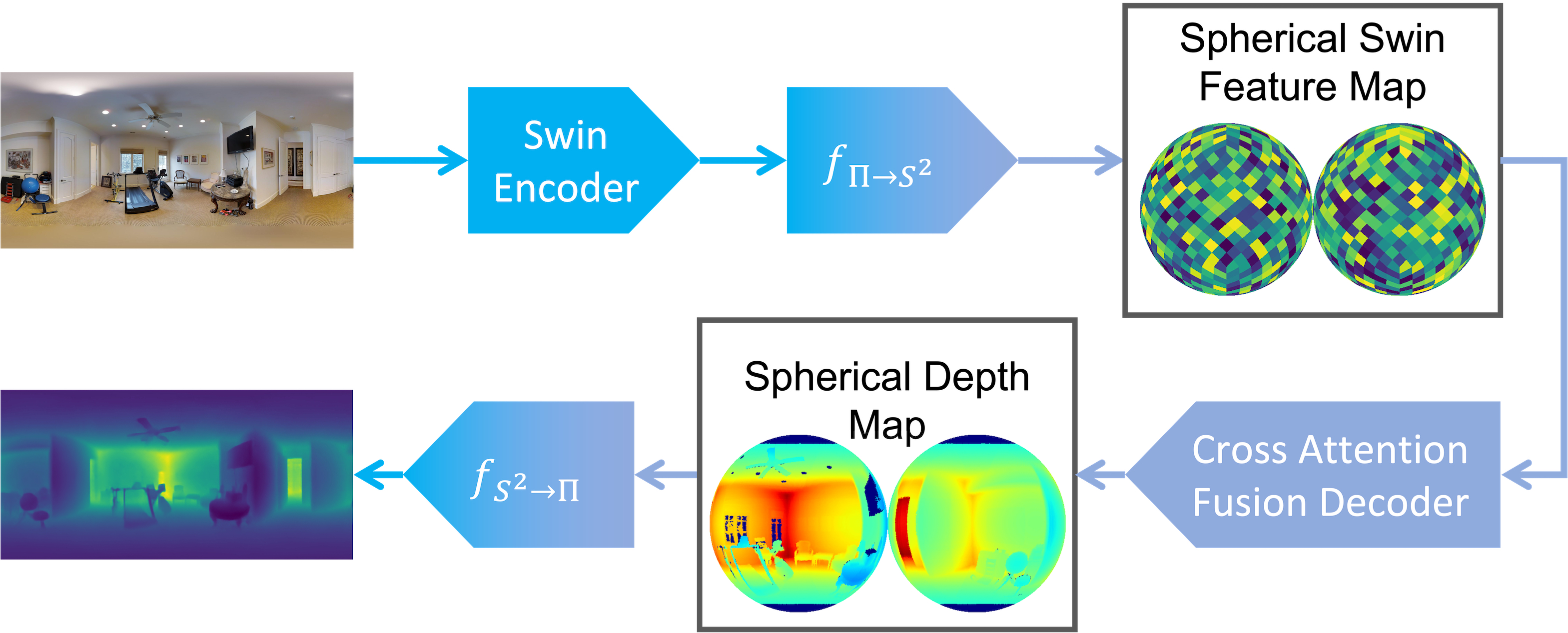}
    \caption[Pipeline of the proposed method.]
    {Pipeline of the proposed method. \textcolor{lightblue}{Light blue} parts work on planar ERP images, 
    while \textcolor{greyblue}{grey blue} parts work on unit spherical surface. 
    $\mathbf{f}_{{\Pi}\rightarrow{\mathbf{S^{2}}}}$ denotes mapping from image plane to unit sphere and $\mathbf{f}_{{\mathbf{S^{2}}}\rightarrow\Pi}$ denotes mapping from unit sphere to image plane.}
    \label{fig:1}
\end{figure}

Besides challenges on extending receptive field, for panorama images, additional challenges arise from its unique \textbf{Equi-Rectangular Projection (ERP)} image representation. 
This introduces severe distortion and networks directly working on ERP images are biased by these distortion towards the poles~\cite{albanis2021pano3d}, as the pixels from ERP images are unevenly distributed on the corresponding spherical surface.
Also, directly learning from ERP image representation cannot fulfill the implicit assumption of panorama images that the left side and the right side of the image are physically connected, 
which results in spatial inconsistency on the depth map~\cite{sun2021indoor}.

Some previous methods~\cite{wang2020bifuse,jiang2021unifuse} employ an additional branch where cubemaps generated from panorama images are introduced to compensate for the distortion. 
However, discontinuity happens on seams between different sides of cubemaps, thus the distortion on the ERP image cannot be fully compensated. 
Other methods~\cite{zhuang2021acdnet,zioulis2018omnidepth,li2022omnifusion} take another way by using irregular convolution kernel or dilated convolution, 
but their distortion removal is limited by the learning ability of the network.
In addition, all of the above methods need circular padding to ensure spatial consistency of the depth map, but the problem cannot be fully solved. 

In a word, these problems require the panorama depth estimation method can represent ERP images in a distortion-reduced manner in which the pixels are evenly distributed, and possesses large receptive field that can better explore embedded global structural information. 

Inspired by previous works on spherical CNNs~\cite{du2021spherical,cohen2018s2cnn},
we move the panorama depth estimation task onto spherical surface generated from ERP image using sampling method HEALPix~\cite{gorski2005healpix},
which guarantees that the sampled pixels distribute uniformly on sphere thus can largely reduces distortion. 
Additional advantages is that the spherical representation naturally guarantees spatial consistency of the two sides of panorama images, 
which reduces the need for remedial methods like circular padding.

We also introduce \textbf{Cross Attention Fusion (CAF)} module that utilizes learnable cross attention to fuse feature maps from skip connection and to enhance the capability of the decoder to retrieve global context information.
The pipeline of the proposed method can be found in Figure~\ref{fig:1}.
To the best of our knowledge, the proposed method is the first one that employs a vision transformer on the unit sphere in real-world vision tasks.

The proposed method outperforms previous methods significantly on nearly all public datasets, 
especially on new Pano3D~\cite{albanis2021pano3d} datasets which focuses on method's generalization capablity: 
the $RMSE$ error decreases $\mathbf{23.3}\%$ and $AbsRel$ error decreases $\mathbf{16.3}\%$.
Our method also obtains good results in spatial consistency and robustness to misalignment on gravity direction. 

\textbf{Main contributions of this paper} are: 
\begin{enumerate}
    \item A novel end-to-end deep network that moves decoder network onto spherical surface 
    which eliminates the distortion of ERP images and maintains the spatial consistency of depth maps.
    \item An spherical-transformer-based fusion module (CAF) that fuses features from skip connection in a learnable manner 
    and enhances the capability of retrieving global context information.
\end{enumerate}

The paper is organized as follows:
Section~\ref{sec:2} reviews and discusses related works;
Section~\ref{sec:3} contains the necessary details of the proposed method;
Section~\ref{sec:4} demonstrates experiments and ablation studies;
Section~\ref{sec:5} concludes the paper.

\section{Related Works}
\label{sec:2}

\paragraph{Depth estimation for perspective camera images} 
new methods use ideas such as knowledge distillation~\cite{wang2021knowledge}, attention-based methods~\cite{bhat2021adabins,chen2021attention} and self-supervision~\cite{watson2021temporal}. 
Extending the global context sensing capability of the network is also a focused research topic.
Methods~\cite{bhat2021adabins,huynh2020guiding} try to extend the global context of their method by adding vision transformer based processing module into the network, but the former work's~\cite{bhat2021adabins} traditional part of network's global context have not been enlarged, and the latter work~\cite{huynh2020guiding} requires additional supervising for the module. 
Previous work~\cite{ranftl2021vision} adds attention-based modules as its feature extraction module, but its decoder remains to be the traditional structure.

\paragraph{Depth estimation for panorama images} 
some methods~\cite{wang2020bifuse,jiang2021unifuse} use cubemaps as the image representation method, 
and try to use information derived from cubemaps to compensate for ERP image's distortion, 
which results may be affected by discontinuity of cubemaps on each side's edges. 
There also exist methods~\cite{zhuang2021acdnet,zioulis2018omnidepth} use specialized convolution methods such as dilated convolution to tackle the distortion on ERP images and enlarge the receptive area of the network. 
These methods do not use distortion-reduced methods to represent ERP images,
and hence results may be affected.
Some methods try to represent ERP images with 1D representations~\cite{pintore2021slicenet,sun2021hohonet}; 
however, distortion on latitude direction may not be fully removed by 1D representation.
Compared to these previous works, the proposed method uses distortion-reduced representation on the sphere, which is not affected by discontinuity from cubemap, and the receptive field of the proposed method can be further enlarged by vision methods.

\begin{figure}
    \centering 
    \includegraphics[width=0.8\linewidth]{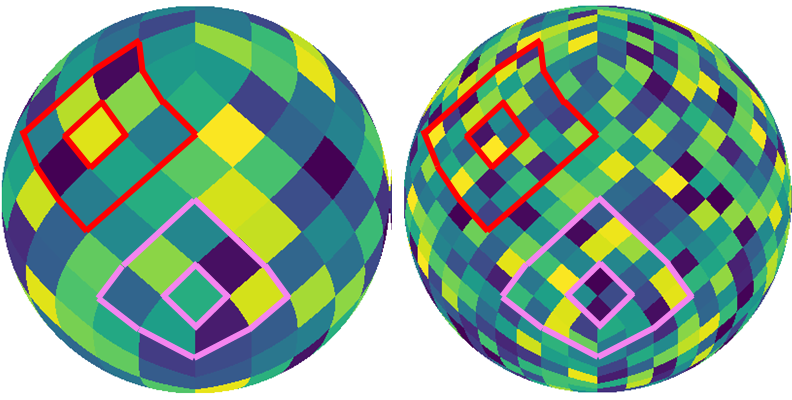}
    \caption[HEALPixel.]
    {We demonstrate the multi-scale sampling capability of HEALPix by two examples with different $n_{size}$: $n_{size}=4$ on the left and $n_{size}=8$ on the right. }
    \label{fig:2}
\end{figure}

\paragraph{Spherical deep learning} most of these methods use dilated convolution or convolution kernel with special forms~\cite{Su2020KTN,fernandez2019CFL}, 
or compensate distortion using latitude and longitude~\cite{Wang2020360SDNET,Liu18CoordConv} in panorama image understanding tasks. 
There are two works that are similar to the proposed method: 
one work~\cite{cho2022spherical} utilizes the Icosahedron sampling method~\cite{Lee2022SpherePHD} and forms a novel transformer for representation learning of ERP images. 
Another work~\cite{du2021spherical} utilizes HEALPix~\cite{gorski2005healpix} as the sampling method and proposes a CNN method for this unique pixel representation. 
Compared to the above-mentioned methods, our method is the first with complete infrastructure for vision transform techniques used on the unit sphere, and the accuracy of panorama depth estimation using this method is significantly improved.

\section{Approach}
\label{sec:3}

We use the classic encoder-decoder deep network architecture with skip-connections. 
Multi-scale feature maps extracted from the input ERP image by a vision transformer encoder network are transitioned from their planar representation to a uniform spherical pixel representation. 
Then the multi-scale decoder network performs depth regression along the entire surface of the sphere to predict depth estimation results. 
The multi-scale decoder network consisted by several fusion modules at different scales on the unit sphere is used to fuse the spherical feature maps from the skip connection which contains fine-grained local details learned in the encoder part.

\subsection{Panorama Image Representation}
\label{sec:3.1}

We use HEALPix (the Hierarchical Equal Area isoLatitude Pixelization)~\cite{gorski2005healpix} to generate uniformly distributed multi-scale pixels on the unit sphere; 
for ${n}_{size}$ samples per-level, 
the total number of sampled pixel would be ${n}_{pix}=12 * {{n}_{size}}^2$.
We use 4 coarse-to-fine spherical levels which pixel numbers are $[192, 768, 3072, 12288]$.
Exampled spherical grids can be found in Figure~\ref{fig:2}.

It is worth noticing that either using fewer (downsample) or more (upsample) pixels in HEALPix will not bring positive effects on final results. Tests can be found in Table~\ref{tab:4.0}.
We upsample and downsample the pixel number to $[768, 3072, 12288, 12288]$ and $[48, 192, 768, 3072]$ respectively, to study its effects on final accuracy results. 
Due to limited GPU memory, we do not upsample the finest level.
The proposed pixel number achieves overall best accuracy results,
the reason is probably two-fold:     
less pixel brings larger area of one particular pixel, which carries more global context information but less details;
more pixel helps the network retrieve details in depth map, but will lose global context information. 
The proposed pixel number seems to be a good bargain between these two factors.

\begin{figure}
    \centering 
    \includegraphics[width=0.93\linewidth]{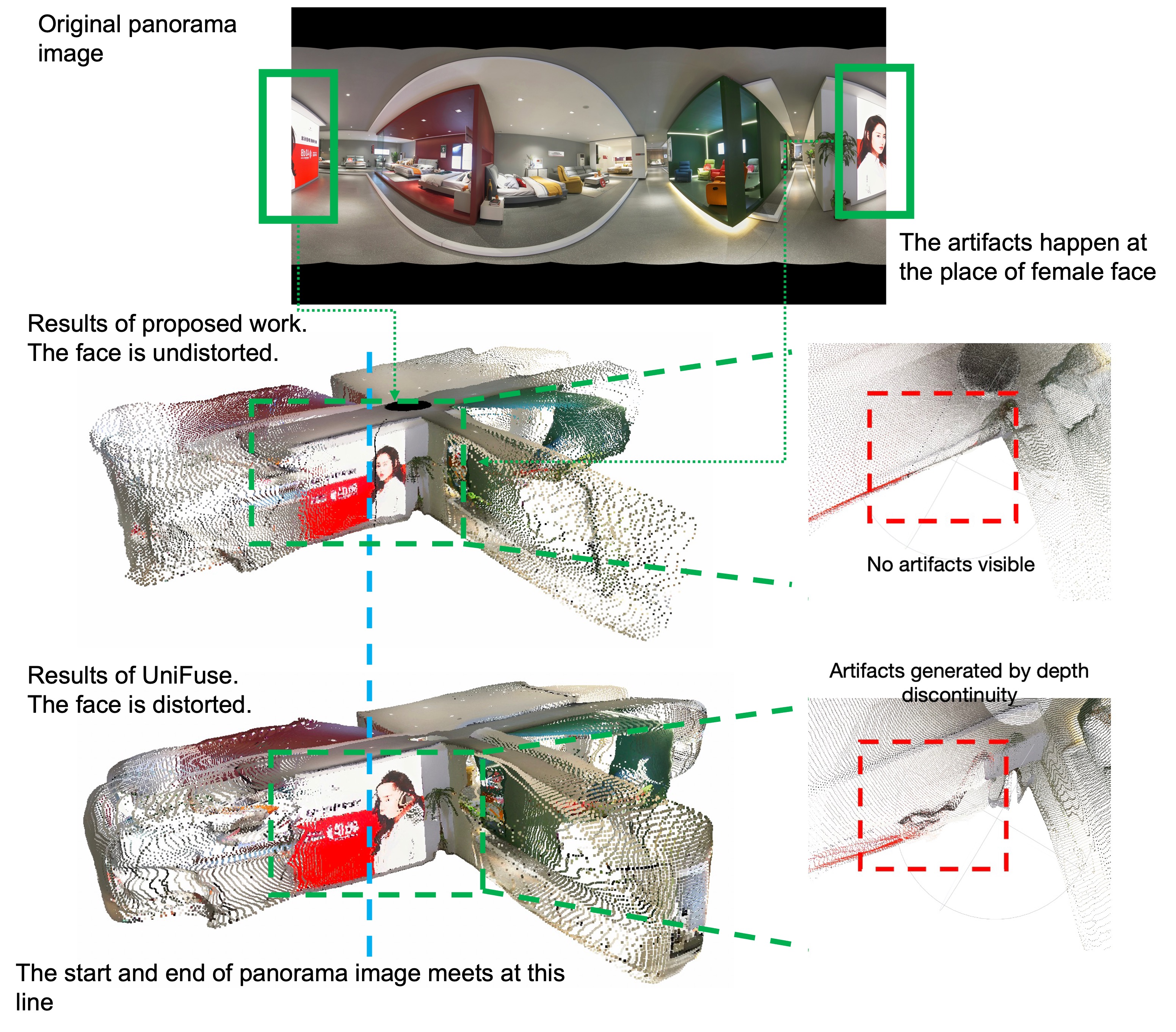}
    \caption[Compare point cloud.]
    {We compare the spatial consistency of UniFuse (\textbf{with circular padding})~\cite{jiang2021unifuse} and the proposed method. Methods like UniFuse cannot guarantee spatial consistency of the left and right end of the ERP images, while the proposed method solves this problem.}
    \label{fig:4}
\end{figure}

\begin{table}
  \centering
  \scriptsize
  \caption{Comparison of effects of different HEALPix pixel numbers on final depth map accuracy on M3D dataset~\cite{chang2017matterport3d}.}
  \label{tab:4.0}   

  \begin{tabular}{c|c|c|c|c}
  \hline\noalign{\smallskip}
  \multirow[c]{2}{*}{Pixel number} & \multicolumn{3}{c|}{Error $\Downarrow$} & Accuracy (\%) $\Uparrow$ \\
  
  & ${AbsRel}$ & ${RMSE}$ & ${RMSLE}$ & $\delta < 1.25$ \\
  \noalign{\smallskip}\hline
  \hline\noalign{\smallskip}
  Downsample & 0.0902 & 0.4211 & 0.0582 & 92.07 \\
  Original & \tablebestresult{0.0865} & \tablebestresult{0.4052} & \tablebestresult{0.0559} & \tablebestresult{92.63} \\
  Upsample & 0.0866 & 0.4109 & 0.0565 & 92.54 \\
  \noalign{\smallskip}\hline
  \end{tabular}
  
\end{table}

HEALPix helps define two mappings between the surface and the image plane: 
the mapping from image plane to unit sphere $\mathbf{f}_{{\Pi}\rightarrow{\mathbf{S^{2}}}}$ and mapping from unit sphere to image plane $\mathbf{f}_{{\mathbf{S^{2}}}\rightarrow\Pi}$, 
which can be pre-computed using open-source software\footnote{\url{https://healpy.readthedocs.io/en/latest/}};
the pre-computed mapping can be used to formulate neighbourhood relation look-up table for each pixel, 
which ensures easy implementation of self-attention calculation.
We use bilinear interpolation to calculate pixel color for each pixel on HEALPix sphere,
using other interpolation methods nearly makes no difference to final accuracy.
The encoder operates on the original ERP image, 
and the generated feature maps with different channels are mapped onto the spherical surface with unified channel $D\in\{256, 320, 512\}$ using HEALPix.
Unit sphere-based image representation does not have boundaries, 
thus operations on the unit sphere is naturally circular, 
which makes the proposed method better at maintain spatial consistency on the left side and right side of ERP images, comparing with methods like UniFuse~\cite{jiang2021unifuse} with circular padding, 
as both the receptive field and the padding’s width are limited. See Figure~\ref{fig:4} for one vivid example.

\begin{figure}
    \centering 
    \includegraphics[width=0.93\linewidth]{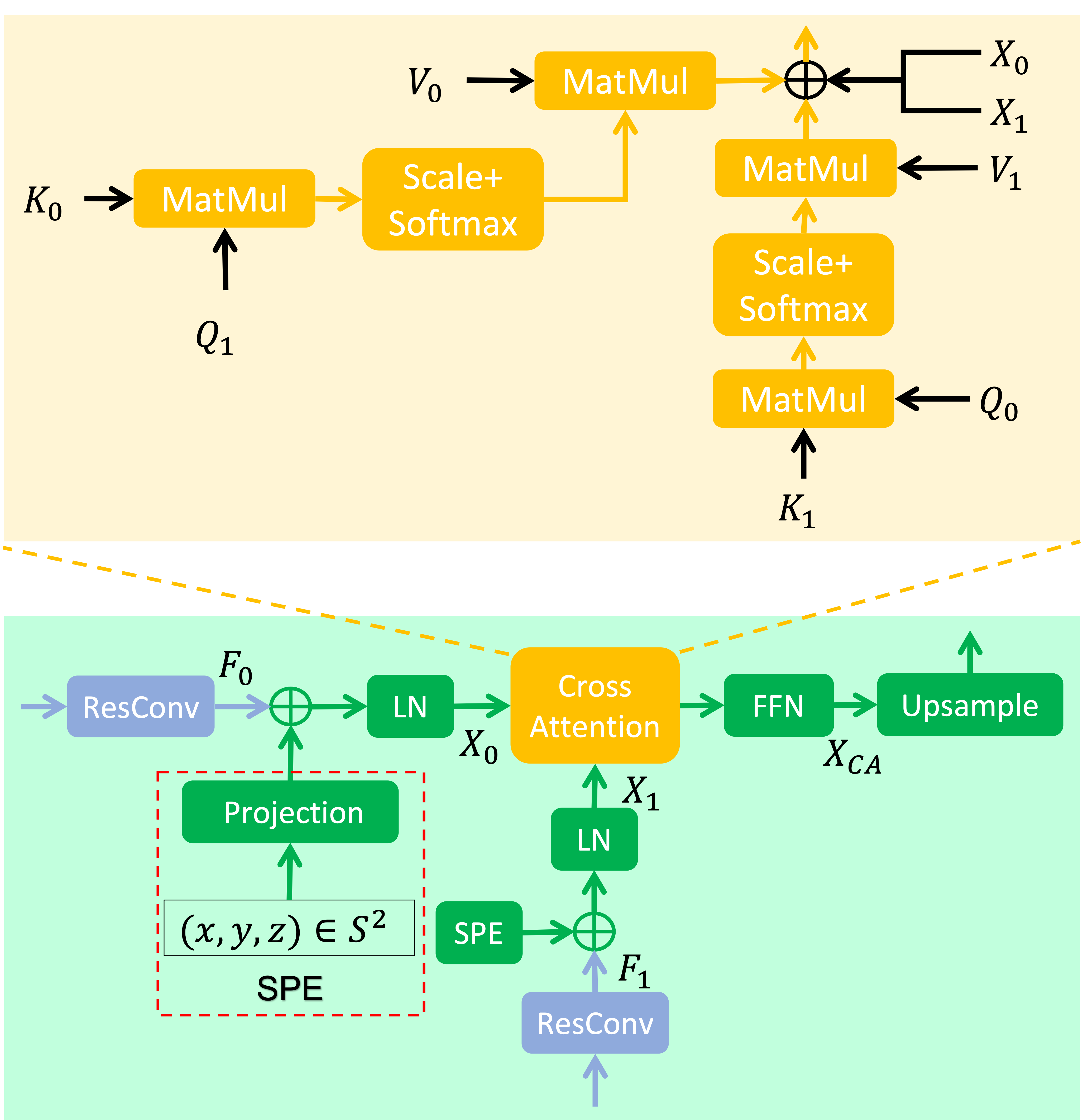}
    \caption[Compare point cloud.]
    {Structure of CAF module that conducts self-attention calculation on sphere.
    ``LN", ``SPE", and ``ResCon" denote Layer Normalization, Spherical Positional Embedding (Section~\ref{sec:3.4}), and Residual Convolution, respectively. }
    \label{fig:4.1}
\end{figure}

\begin{figure*}
    \centering 

    \includegraphics[width=0.85\linewidth]{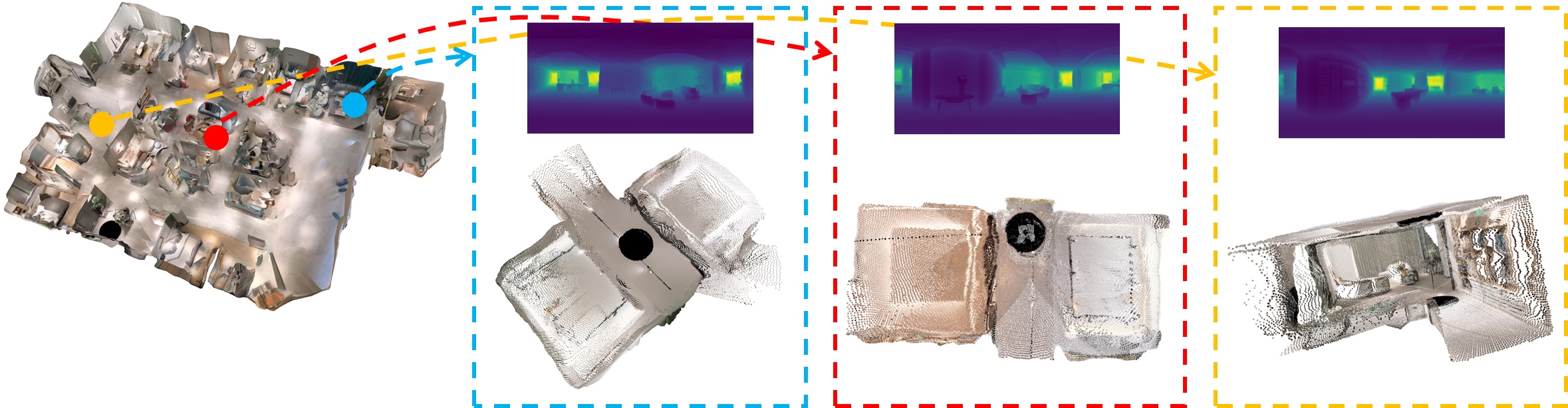}
    \caption[Result of depth map generated.]
    {We collect a series of panorama images at different spots in a real-world scene, 
    and estimate a depth map for each image. 
    These depth maps are then registered to one global coordinate system by Structure from Motion.
    We demonstrate the final mesh model generated from alignment results on the left side of the figure,
    and depth maps for three exampled spots in the scene and
    corresponding point clouds on the right side. }
    \label{fig:3}
\end{figure*}

\subsection{Cross Attention Fusion}
\label{sec:3.3}

Multiple \textbf{Cross Attention Fusion (CAF)} modules that operates on multi-scale spherical features formulate the multi-scale decoder. 
CAF is designed to enhance the network's global context acquisition capability and to fuse feature maps from skip connection to enhance the details of estimated depth map. 

CAF adopts a ``global" cross attention calculation mechanism, which means that correlations are computed among pixels of feature maps at different positions globally. 
In this way, CAF has the ability of long-dependencies and can learn better global context information from multi-path feature maps. Features from skip-connections contain ``local" fine-grained details as they come from pyramidal encoder network, 
while the features from the decoder are relatively ``global" and possess larger receptive field. 
Previous global-local fusion strategies mostly adopt trivial sum or concatenate, 
which directly mixes up these feature maps with different contexts. 
Our CAF module uses a learnable global attention based fusion strategy: 
feature maps from skip connection with local context are treated as ``guidance" to enhance feature maps from decoder with global context, and vice versa.
In this way, CAF enhances the correlation between two feature maps by learning compensation values. The detailed structure of the CAF module can be found in Figure~\ref{fig:4.1}.

We use \textit{Base} to denote fusion module with direct addition, \textit{Fusion} to denote fusion module with our CAF. The \textit{Base} fusion module is only consisted of two simple residual convolution units from work~\cite{ranftl2021vision} using basic spherical convolution method in previous works~\cite{du2021spherical,cho2022spherical}.

\begin{figure*}
    \centering 

    \includegraphics[width=0.85\linewidth]{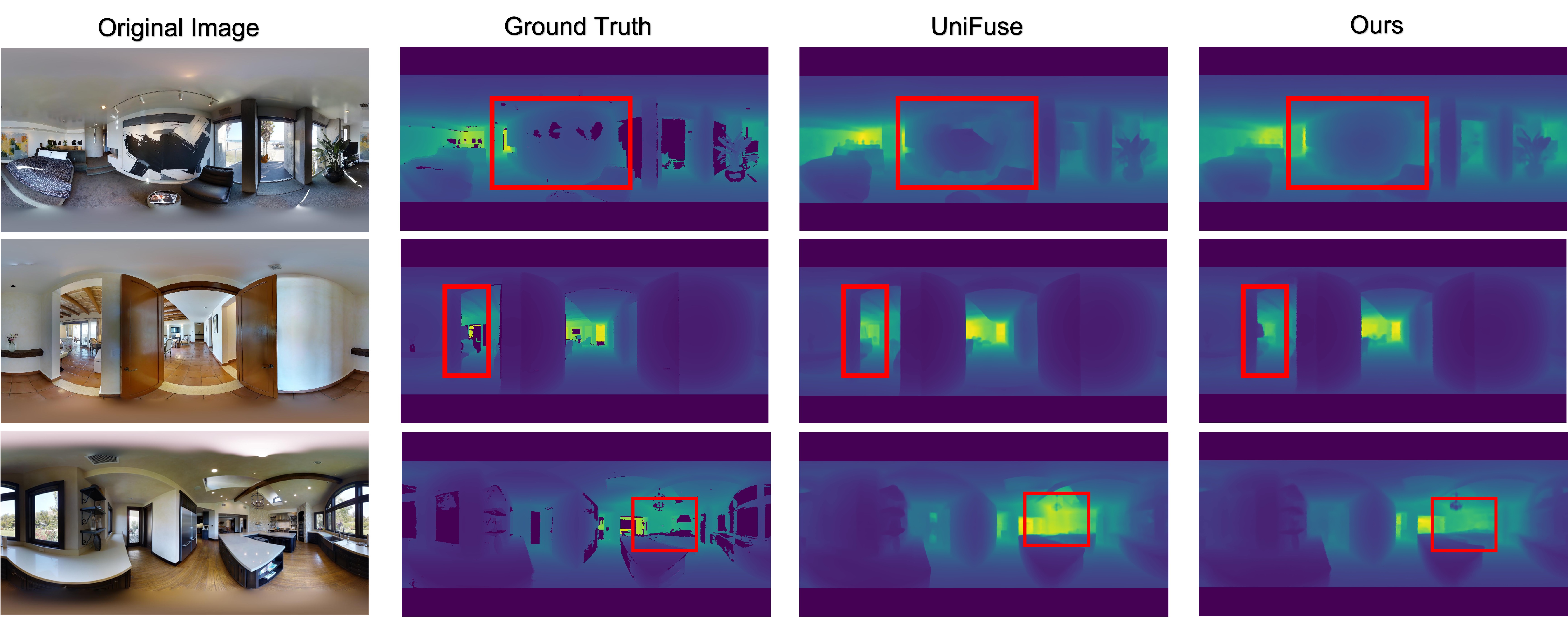}
    \caption[Result of depth map generated.]
    {Comparison between the proposed work and UniFuse~\cite{jiang2021unifuse} on several selected scenes from the M3D dataset. We use boxes to mark places worth noticing. Note that we use different colors to distinguish different noticeable places.}
    \label{fig:3.1}
\end{figure*}

\begin{table}
  \centering
  \scriptsize
  \caption{Comparison of effects of formulation of coordinates used in SPE on final depth map accuracy on M3D dataset~\cite{chang2017matterport3d}. 
  We use $[lat,lon]$ to represent SPE using point's longitude and latitude information.}
  \label{tab:4.3}   

  \begin{tabular}{c|c|c|c|c}
  \hline\noalign{\smallskip}
  \multirow[c]{2}{*}{Coordinates} & \multicolumn{3}{c|}{Error $\Downarrow$} & Accuracy (\%) $\Uparrow$ \\
  & ${AbsRel}$ & ${RMSE}$ & ${RMSLE}$ & $\delta < 1.25$ \\
  \noalign{\smallskip}\hline
  \hline\noalign{\smallskip}
  $[x,y]$ & 0.0910 & 0.4238 & 0.0589 & 91.83 \\
  $[y,z]$ & 0.0900 & 0.4202 & 0.0581 & 92.21 \\
  $[x,z]$ & 0.0874 & 0.4162 & 0.0570 & 92.60 \\
  $[lat,lon]$ & 0.0954 & 0.4416 & 0.0616 & 90.92 \\
  $[x,y,z]$ & \tablebestresult{0.0865} & \tablebestresult{0.4052} & \tablebestresult{0.0559} & \tablebestresult{92.63} \\
  \noalign{\smallskip}\hline
  \end{tabular}
  
\end{table}

\begin{table}
  \centering
  \scriptsize
  \caption{Comparison of effects of position embedding methods on final depth map accuracy on M3D dataset~\cite{chang2017matterport3d}}
  \label{tab:6}   

  \begin{tabular}{c|c|c|c|c}
  \hline\noalign{\smallskip}
  \multirow[c]{2}{*}{Encoding method} & \multicolumn{3}{c|}{Error $\Downarrow$} & Accuracy (\%) $\Uparrow$ \\
  & ${AbsRel}$ & ${RMSE}$ & ${RMSLE}$ & $\delta < 1.25$ \\
  \noalign{\smallskip}\hline
  \hline\noalign{\smallskip}
  Learnable & 0.0881 & 0.4142 & 0.0570 & 92.369 \\
  SPE & \tablebestresult{0.0865} & \tablebestresult{0.4052} & \tablebestresult{0.0559} & \tablebestresult{92.634} \\
  \noalign{\smallskip}\hline
  \end{tabular}
  
\end{table}

Given feature map $F_{i}$ from skip-connection and decoder  network and positional embedding ${\zeta}_{i}$, the input of CAF can be defined as:
\begin{equation}
\begin{aligned}
X_{i} = LN(F_{i} + {\zeta}_{i}), i \in \{0, 1\} 
\end{aligned}
\label{eq:1}
\end{equation} 
CAF module tries to estimate two compensation terms ${Att}_{0}$ and ${Att}_{1}$ by cross-attention: 
\begin{equation}
\begin{aligned}
{{Att}_{0}}(Q_{1}, K_{0}, V_{0}) &= {SoftMax}(Q_{1}K_{0} / \sqrt{d})V_{0} \\
{{Att}_{1}}(Q_{0}, K_{1}, V_{1}) &= {SoftMax}(Q_{0}K_{1} / \sqrt{d})V_{1} 
\end{aligned}
\label{eq:2}
\end{equation} 
the two values are added into inputs:
\begin{equation}
\begin{aligned}
X_{CA} &= {FFN}({LN}(X_{0} + {Att}_{0} + X_{1} + {Att}_{1}))  \\
\end{aligned}
\label{eq:3}
\end{equation} 
$FFN$ denotes feed-forward operation, and $LN$ denotes layer norm operation~\cite{cho2022spherical}, the two operations formulate a residual-like structure. 
$Q_{i}$, $K_{i}$ and $V_{i}$ used in Equation~\ref{eq:2} are generated from input ${X}_{0}$ and ${X}_{1}$ using corresponding learnable parameters $W$: 
\begin{equation}
\begin{aligned}
Q_{i} = W^{Q}X_{i}, K_{i} = W^{K}X_{i}, V_{i} = W^{V}X_{i}, i \in \{0, 1\} 
\end{aligned}
\label{eq:4}
\end{equation} 

As the global attention calculation requires large computing resources, we adopt the method used in previous work~\cite{chu2021twins} to perform global sub-sampled cross attention utilizing non-overlapped window partitions. 
These partitions can be conveniently derived from the coarse-to-fine structure of HEALPix.

\begin{table}
  \centering
  \scriptsize
  \caption{Comparison of effects of different loss function on final depth map accuracy on M3D datasets~\cite{chang2017matterport3d}. 
  We use ``RMSERel'' to represent Relative RMSE~\cite{Alvarez-Gila2017Adversarial}.}
  \label{tab:4.1}   

  \begin{tabular}{c|c|c|c|c}
  \hline\noalign{\smallskip}
  \multirow[c]{2}{*}{Loss Function} & \multicolumn{3}{c|}{Error $\Downarrow$} & Accuracy (\%) $\Uparrow$ \\
  & ${AbsRel}$ & ${RMSE}$ & ${RMSLE}$ & $\delta < 1.25$ \\
  \noalign{\smallskip}\hline
  \hline\noalign{\smallskip}
  RMSERel & 0.0943 & 0.4200 & 0.0597 & 91.49 \\
  RMSLE & \tablebestresult{0.0865} & \tablebestresult{0.4052} & \tablebestresult{0.0559} & \tablebestresult{92.63} \\
  \noalign{\smallskip}\hline
  \end{tabular}
  
\end{table}

\subsection{Spherical Positional Embedding}
\label{sec:3.4}

In order to make the CAF module better obtains pixel position information, we proposed a handcrafted positional embedding method, namely \textbf{Spherical Positional Embedding (SPE)}.
We concatenate the spherical coordinates along the channel axis, and then embed the three-channel input to the dimension of the feature maps: 
\begin{equation}
\begin{aligned}
\zeta = C\times{[x_{sphere},y_{sphere},z_{sphere}]}^{T}
\end{aligned}
\label{eq:4}
\end{equation}
where $C\in\mathbb{R}^{M\times3}$ is an embedding matrix, $M$ is the channels of the feature map.

Reducing the dimensions of coordinate being used, or using latitude and longitude to form positional embedding, 
will reduce the overall accuracy of the proposed method;
corresponding tests can be found in Table~\ref{tab:4.3}.
Large accuracy decrease when using latitude and longitude is probably because same area of pixel produce different changes in longitude when the pixel is placed on different latitude,
which may introduce implicit distortion into final results.

Compared with the learnable positional embedding mechanism, this embedding method utilizes the real space location on the spherical surface, 
which helps the decoder network and CAF module take into account non-Euclidean spatial information embedded on the sphere, yielding better results.
Results in Table~\ref{tab:6} shows comparison between the two kinds of embedding methods, and the proposed SPE method produces better results, which $RMSE$ is decreased by $\mathbf{2.2}\%$.

\begin{table*}
  \centering
  \scriptsize
  \caption{Comparison of our method with currently available methods on M3D datasets~\cite{chang2017matterport3d} and S2D3D datasets~\cite{armeni2017jointstanford2d3d}.
  We use $^*$ to label results evaluated after performing scale alignment by medium or maximum depth.
  ${\triangle}$ denotes methods that does not contain optimization against distortion on panorama images.
  Note that we use error and accuracy results of SliceNet~\cite{pintore2021slicenet} from ACDNet's paper~\cite{zhuang2021acdnet} in this table.}
  \label{tab:1}   

  \begin{tabular}{c|c|c|c|c|c|c|c|c}
  \hline\noalign{\smallskip}
  \multirow[c]{2}{*}{Dataset} & \multirow[c]{2}{*}{Method} & \multicolumn{4}{c|}{Error Metric $\Downarrow$} & \multicolumn{3}{c}{Accuracy Metric (\%) $\Uparrow$} \\
  & & ${MAE}$ & ${AbsRel}$ & ${RMSE}$ & ${RMSLE}$ & $\delta < 1.25$ & $\delta < {1.25}^{2}$ & $\delta < {1.25}^{3}$ \\
  \noalign{\smallskip}\hline
  \hline\noalign{\smallskip}
 \multirow[c]{11}[3]{*}{M3D} & FCRN$^*$~\cite{laina2016fcrn} & 0.4008 & 0.2409 & 0.6704 & 0.1244 & 77.03 & 91.74 & 96.17 \\
  & SliceNet$^*$~\cite{pintore2021slicenet} & 0.3296 & 0.1764 & 0.6133 & 0.1045 & 87.16 & 94.83 & 97.16 \\
  & \textbf{Ours (base)}$^*$ & 0.2025 & 0.0776 & 0.3787 & 0.0513 & 93.58 & 98.31 & 99.44  \\
  & \textbf{Ours (fusion)}$^*$ & \aligntablebestresult{0.1932} & \aligntablebestresult{0.0730} & \aligntablebestresult{0.3720} & \aligntablebestresult{0.0498} & \aligntablebestresult{93.96} & \aligntablebestresult{98.38} & \aligntablebestresult{99.46}  \\
  \cmidrule[1.5pt]{2-9}
  & Bifuse~\cite{wang2020bifuse} & 0.3470 & 0.2048 & 0.6259 & 0.1134 & 84.52 & 93.19 & 96.32  \\
  & HoHoNet~\cite{sun2021hohonet} & 0.2862 & 0.1488 & 0.5138 & 0.0871 & 87.86 & 95.19 & 97.71  \\
  & UniFuse~\cite{jiang2021unifuse} & 0.2814 & 0.1063 & 0.4941 & 0.0701 & 88.97 & 96.23 & 98.31  \\
  & ACDNet~\cite{zhuang2021acdnet} & 0.2670 & 0.1010 & 0.4629 & 0.0646 & 90.00 & 96.78 & 98.76  \\
  & OmniFusion~\cite{li2022omnifusion} & $\times$ & 0.0900 & 0.4261 & 0.1483 & 91.89 & \tablebestresult{97.97} & \tablebestresult{99.31}  \\
  & \textbf{Ours (base)} & 0.2346 & 0.0906 & 0.4153 & 0.0577 & 91.92 & 97.55 & 99.12  \\
  & \textbf{Ours (fusion)} & \tablebestresult{0.2263} & \tablebestresult{0.0865} & \tablebestresult{0.4052} & \tablebestresult{0.0559} & \tablebestresult{92.64} & 97.68 & 99.11  \\
  \cmidrule[1.5pt]{2-9}
  & ${\triangle}$AdaBins~\cite{bhat2021adabins} & 0.2560 & 0.0958 & 0.4484 & 0.1426 & 90.54 & 97.05 & 98.92  \\
  \midrule[1.5pt]
  \multirow[c]{11}[3]{*}{S2D3D} & FCRN$^*$~\cite{laina2016fcrn} & 0.3428 & 0.1837 & 0.5774 & 0.1100 & 72.30 & 92.07 & 97.31 \\
  & SliceNet$^*$~\cite{pintore2021slicenet} & 0.1757 & 0.0995 & 0.3509 & 0.0801 & 90.29 & 96.26 & 98.44  \\
  & \textbf{Ours (base)}$^*$ & 0.1745 & 0.0919 & \aligntablebestresult{0.3229} & 0.0593 & 92.12 & \aligntablebestresult{98.14} & \aligntablebestresult{99.31}  \\
  & \textbf{Ours (fusion)}$^*$ & \aligntablebestresult{0.1655} & \aligntablebestresult{0.0843} & 0.3232 & \aligntablebestresult{0.0575} & \aligntablebestresult{92.90} & 98.09 & 99.29  \\
  \cmidrule[1.5pt]{2-9}
  & Bifuse~\cite{wang2020bifuse} & 0.2343 & 0.1209 & 0.4142 & 0.0787 & 86.60 & 95.80 & 98.60  \\
  & HoHoNet~\cite{sun2021hohonet} & 0.2027 & 0.1014 & 0.3834 & 0.0668 & 90.54 & 96.93 & 98.86  \\
  & UniFuse~\cite{jiang2021unifuse} & 0.2082 & 0.1114 & 0.3691 & 0.0721 & 88.97 & 96.23 & 98.31  \\
  & ACDNet~\cite{zhuang2021acdnet} & 0.1870 & 0.0984 & 0.3410 & 0.0664 & 88.72 & 97.04 & 98.95  \\
  & OmniFusion~\cite{li2022omnifusion} & $\times$ & 0.0950 & 0.3474 & 0.1599 & 89.88 & 97.69 & \tablebestresult{99.24}  \\
  & \textbf{Ours (base)} & \tablebestresult{0.1760} & 0.0936 & \tablebestresult{0.3272} & \tablebestresult{0.0600} & 91.53 & \tablebestresult{98.30} & 99.18 \\
   & \textbf{Ours (fusion)} & 0.1774 & \tablebestresult{0.0903} & 0.3383 & \tablebestresult{0.0600} & \tablebestresult{91.91} & 97.82 & 99.12  \\
  \noalign{\smallskip}\hline
  \end{tabular}
  
\end{table*}

\subsection{Training Loss}
\label{sec:3.5}

We use Root Mean Squared Logarithmic Error (RMSLE) as the loss function to supervise the training process, as shown in Equation~\ref{eq:7}. 
$d_{i}$ and $\overline{d_{i}}$ refer to estimated and ground truth depth for pixel $i$ respectively. 
\begin{equation}
 \begin{aligned}
    \mathcal{L}=\sqrt{\frac{1}{N}\sum_{i=1}^{N}(\log(d_{i})-\log(\overline{d_{i}}))^2}
\end{aligned}
\label{eq:7}
\end{equation}
We optimize for network parameters on all spherical pixels with valid depth value. 
RMSLE can reduce the impact of large differences between prediction and ground truth on the overall loss and ensure that the loss is uniformly distributed in different depth ranges.
As the spherical pixels are already distributed uniformly on sphere, 
the contribution of each pixel to the overall loss is spatially uniform. 
We also notice that \textbf{Relative RMSE (RMSERel)}~\cite{Alvarez-Gila2017Adversarial} can also reduce the impact of large error;
we compare the accuracy results of the two kinds of losses in Table~\ref{tab:4.1}.
Results of RMSERel is worse than proposed RMSLE, 
this is probably because RMSLE suppress losses on faraway objects more than RMSERel, 
resulting in better spatial measurements on panorama depth estimation tasks, 
as most of these tasks focus on indoor \ie near scenarios.

\section{Experiments}
\label{sec:4}

\begin{table*}
  \centering
  \scriptsize
  \caption{Results on Pano3D~\cite{albanis2021pano3d} datasets.
  ``$w$'' means the error uniformly distribute by re-weighting using weighting methods provided by Pano3D.
  Metrics without Pano3D official results are marked with "$\times$".}
  \label{tab:3}   

  \begin{tabular}{c|c|c|c|c|c|c|c|c|c}
  \hline\noalign{\smallskip}
  \multirow[c]{2}[3]{*}{Dataset} &\multirow[c]{2}[3]{*}{Method} & \multicolumn{4}{c|}{Error Metric $\Downarrow$} & \multicolumn{4}{c}{Accuracy Metric (\%) $\Uparrow$} \\
  & & ${\mathit{w}RMSE}$ & ${\mathit{w}RMSLE}$ & ${\mathit{w}AbsRel}$ & ${\mathit{w}SqRel}$ & ${\delta}_{1.1}^{{ico}^6}$ & ${\delta}_{1.25}^{{ico}^6}$ & ${\delta}_{{1.25}^{2}}^{{ico}^6}$ & ${\delta}_{{1.25}^{3}}^{{ico}^6}$ \\
  \noalign{\smallskip}\hline
  \hline\noalign{\smallskip}
  \multirow[c]{4}{*}{\tabincell{c}{\textit{train on M3D} \\ test on GV2 tiny}} & ${\mathrm{UNet}}^{vnl}$ & 0.5794 & 0.1247 & 0.2151 & $\times$ & 31.98 & 62.05 & $\times$ & $\times$ \\
  & ${\mathrm{ResNet}}^{comp}_{skip}$ & 0.4993 & 0.1273 & 0.1758 & $\times$ & 40.78 & 80.31 & $\times$ & $\times$ \\
  & \textbf{Ours (base)} & \tablebestresult{0.3654} & 0.0800 & 0.1458 & 0.0675 & 45.44 & 80.93 & 96.03 & 99.13  \\
  & \textbf{Ours (fusion)} & 0.3682 & \tablebestresult{0.0722} & \tablebestresult{0.1353} & \tablebestresult{0.0615} & \tablebestresult{47.82} & \tablebestresult{85.23} & \tablebestresult{97.45} & \tablebestresult{99.43}  \\
  \cmidrule[1.3pt]{1-10}
  \multirow[c]{4}{*}{\tabincell{c}{\textit{train on M3D} \\ test on GV2 medium}} & ${\mathrm{UNet}}^{vnl}$ & 0.5901 & 0.1291 & 0.2269 & $\times$ & 31.21 & 61.02 & $\times$ & $\times$ \\
  & ${\mathrm{ResNet}}^{comp}_{skip}$ & 0.4528 & 0.1618 & 0.1664 & $\times$ & 42.03 & 81.91 & $\times$ & $\times$ \\
  & \textbf{Ours (base)} & 0.3502 & 0.0806 & 0.1500 & 0.0684 & 44.78 & 80.67 & 95.72 & 99.00  \\
  & \textbf{Ours (fusion)} & \tablebestresult{0.3472} & \tablebestresult{0.0729} & \tablebestresult{0.1393} & \tablebestresult{0.0614} & \tablebestresult{46.95} & \tablebestresult{84.66} & \tablebestresult{97.16} & \tablebestresult{99.33} \\
  \cmidrule[1.3pt]{1-10}
  \multirow[c]{4}{*}{\tabincell{c}{\textit{train on M3D} \\ test on GV2 fullplus}} & ${\mathrm{UNet}}^{vnl}$ & 0.8772 & 0.1769 & 0.2730 & $\times$ & 22.46 & 46.09 & $\times$ & $\times$ \\
  & ${\mathrm{ResNet}}^{comp}_{skip}$ & 0.6607 & 0.2308 & 0.1836 & $\times$ & \tablebestresult{41.18} & \tablebestresult{74.77} & $\times$ & $\times$ \\
  & \textbf{Ours (base)} & 0.5560 & 0.1255 & 0.2000 & 0.1430 & 33.27 & 62.20 & 84.66 & 94.05  \\
  & \textbf{Ours (fusion)} & \tablebestresult{0.5243} & \tablebestresult{0.1106} & \tablebestresult{0.1804} & \tablebestresult{0.1264} & 37.35 & 67.99 & \tablebestresult{88.47} & \tablebestresult{95.66} \\
    \cmidrule[2.0pt]{1-10}
  \multirow[c]{3}{*}{\tabincell{c}{\textit{train on GV2 medium} \\ test on GV2 fullplus}} & OmniFusion~\cite{li2022omnifusion} & 0.6159 & 0.1054 & 0.1618 & 0.1418 & 42.87 & 73.39 & 91.36 & 96.40 \\
  & UniFuse~\cite{jiang2021unifuse} & 0.4498 & 0.0894 & 0.1434 & 0.1110 & 48.11 & 79.47 & 93.89 & 97.74 \\
  & \textbf{Ours (fusion)} & \tablebestresult{0.4479} & \tablebestresult{0.0740} & \tablebestresult{0.1318} & \tablebestresult{0.0764} & \tablebestresult{51.25} & \tablebestresult{84.49} & \tablebestresult{97.00} & \tablebestresult{99.03} \\
  \noalign{\smallskip}\hline
  \end{tabular}
  
\end{table*}

\subsection{Datasets}

\label{sec:4.01}

\textbf{Matterport3D (M3D)}~\cite{chang2017matterport3d} and \textbf{Stanford2D3D (S2D3D)}~\cite{armeni2017jointstanford2d3d} are both real-world datasets collected using Matterport 3D Camera\footnote{\url{https://matterport.com/cameras}}, 
while \textbf{PanoSUNCG (PSunCG)}~\cite{wang2018self} and \textbf{3D60}~\cite{zioulis2018omnidepth} are both synthetic datasets. 
Results evaluated on the above four datasets use metrics from previous works ~\cite{jiang2021unifuse,pintore2021slicenet,sun2021hohonet}, 
including mean absolute error ($MAE$), root mean square error ($RMSE$), logarithmic root mean square error ($RMSLE$), absolute relative error ($AbsRel$) and three threshold percentage~$\delta < {\alpha}^{n}~(\alpha=1.25, n={1, 2, 3})$. 

\textbf{Pano3D}~\cite{albanis2021pano3d} is an evaluation benchmark for panorama depth estimation with the capability of evaluating the network's generalization ability. 
We perform an evaluation on $1024 \times 512$ resolution compared with the baseline methods ${\mathrm{UNet}}^{vnl}$ and ${\mathrm{ResNet}}^{comp}_{skip}$ provided by Pano3D\footnote{\url{https://github.com/VCL3D/Pano3D}}. 
\textbf{Note that metrics with prefix $w$ from evaluations on Pano3D datasets have considered error uniformity problems.}

In Figure~\ref{fig:3}, we demonstrate an aligned mesh generated by multiple depth maps estimated by the proposed method.
Comparison of depth map generated by UniFuse~\cite{jiang2021unifuse} and the proposed method on selected scenes from the M3D datasets are shown in Figure~\ref{fig:3.1}, 
which shows that the proposed method can generate depth maps with better details and is less fragile to image areas with special textures.

\subsection{Implementation Details}
\label{sec:4.02}

The Swin-B~\cite{liu2021swin} pertained on ImageNet-22k is used as the encoder network. 
We use the same data augmentation techniques from UniFuse~\cite{jiang2021unifuse}.
All models are trained on 8 V100 GPU with batch size 32.
We use AdamW~\cite{loshchilov2017decoupled} optimizer with learning rate $\gamma=0.0003$. 
A step decay learning rate scheduler with an interval epoch of 3 and a decay rate of 0.9 is used to accelerate the training process. 
A five epochs warm-up with a starting learning rate of 0.00006 is adopted to get a more stable training convergence. 

\begin{table}

  \centering
  \tiny
  \caption{Comparison of different backbones on final accuracy using M3D datasets~\cite{chang2017matterport3d}. 
  We use EffNet B5 to represent EffcientNet B5 used in AdaBins~\cite{bhat2021adabins},  
  and ``D'' to represent the dimension of decoder. 
  Latency is tested on one NVIDIA TESLA P100 16GB. 
  Planar methods are marked with ${\triangle}$.}
  \label{tab:4}   

  \begin{tabular}{c|c|c|c|c|c|c}
  \hline
  \noalign{\smallskip}
  Method & Backbone & Params & Latency & AbsRel & RMSE & $\delta < 1.25$ \\
  \noalign{\smallskip}\hline\noalign{\smallskip}
  UniFuse~\cite{jiang2021unifuse} & ResNet18 & 30.3M & 25.5ms & 0.1063 & 0.4941 & 88.97 \\
    Base & ResNet18 & \tablebestresult{19.9M} & \tablebestresult{17.3ms} & 0.1061 & 0.4761 & 88.67 \\
  Fusion & ResNet18 & 33.9M & 25.2ms & \tablebestresult{0.0995} & \tablebestresult{0.4614} & \tablebestresult{90.31} \\
  \noalign{\smallskip}\hline\noalign{\smallskip}
   Base & ResNet34 & \tablebestresult{29.9M} & \tablebestresult{47.1ms} & 0.1040 & \tablebestresult{0.4680} & 89.33 \\
  Fusion & ResNet34 & 43.0M & 81.3ms & \tablebestresult{0.0988} & 0.4720 & \tablebestresult{89.97} \\
  \noalign{\smallskip}\hline\noalign{\smallskip}
   ACDNet~\cite{zhuang2021acdnet} & ResNet50 & 87.0M & 72.4ms & 0.1010 & 0.4629 & 90.00 \\
  Base D=256 & ResNet50 & \tablebestresult{32.9M} & \tablebestresult{58.6ms} & 0.1023 & 0.4638 & 89.43 \\
  Base D=400 & ResNet50 & 45.6M & 92.9ms & 0.0999 & 0.4607 & 89.85 \\
  Fusion D=256 & ResNet50 & 45.3M & 99.6ms & 0.0978 & 0.4552 & 90.44 \\
  Fusion D=512 & ResNet50 & 88.7M & 144.3ms & \tablebestresult{0.0944} & \tablebestresult{0.4448} & \tablebestresult{91.07} \\
  \noalign{\smallskip}\hline\noalign{\smallskip}
  ${\triangle}$Adabins~\cite{bhat2021adabins} & EffNet B5 & 78.0M & 199.9ms & 0.0958 & 0.4484 & 90.54 \\
    Base & EffNet B5 & \tablebestresult{39.2M} & \tablebestresult{80.7ms} & 0.0947 & 0.4412 & 91.11 \\
  Fusion & EffNet B5 & 53.2M & 111.2ms & \tablebestresult{0.0911} & \tablebestresult{0.4280} & \tablebestresult{91.90} \\
  \noalign{\smallskip}\hline
  \end{tabular}
\end{table}

\subsection{Overall Accuracy}
\label{sec:4.1}
On real-world datasets, the proposed method achieves overall best performance. As shown in Table~\ref{tab:1}, the decrease on $AbsRel$ of the proposed method is $\mathbf{5.0}\%$ (\textit{Fusion}) on S2D3D and $\mathbf{3.9}\%$ (\textit{Fusion}) on M3D; and the decrease on $RMSE$ is $\mathbf{4.2}\%$ (\textit{Base}) and $\mathbf{4.9}\%$ (\textit{Fusion}) respectively. 
We also compare the proposed method with planar image method AdaBins~\cite{bhat2021adabins}; 
the results of the proposed method is better as it is better in dealing with distortion and retrieving global features.
It's worth noticing that AdaBins provide better results on M3D datasets comparing with ACDNet, 
as self-attention module used in AdaBins largely enhances its capability of extracting features with global context.
\textit{Fusion} method does not outperform \textit{Base} on all evaluation metrics on S2D3D datasets,
as small number of training images (1040 only) in S2D3D makes the CAF module insufficiently trained.

\begin{table*}
  \centering
  \tiny
  \caption{Comparison of our method with currently available methods on synthetic datasets. 
  We use $^*$ to label results evaluated using align methods.
  Most methods trained using the PSunCG datasets can only estimate depth map at resolution $512 \times 256$, 
  but the proposed method can generate higher resolution ($1024 \times 512$), 
  where results are marked with ${\Diamond}$.}
  \label{tab:2}   
  \begin{tabular}{c|c|c|c|c|c|c|c|c}
  \hline\noalign{\smallskip}
  \multirow[c]{2}[3]{*}{Dataset} & \multirow[c]{2}[3]{*}{Method} & \multicolumn{4}{c|}{Error Metric $\Downarrow$} & \multicolumn{3}{c}{Accuracy Metric $\Uparrow$} \\
  & & ${MAE}$ & ${AbsRel}$ & ${RMSE}$ & ${RMSE}_{log}$ & $\delta < 1.25$ & $\delta < {1.25}^{2}$ & $\delta < {1.25}^{3}$ \\
  \noalign{\smallskip}\hline
  \hline\noalign{\smallskip}
  \multirow[c]{11}[6]{*}{PSunCG} & OmniDepth$^*$~\cite{zioulis2018omnidepth} & 0.1624 & 0.1143 & 0.3710 & 0.0882 & 87.05 & 93.65 & 96.50 \\
   & FCRN$^*$~\cite{laina2016fcrn} & 0.1346 & 0.0979 & 0.3973 & 0.0692 & 92.23 & 96.59 & 98.19 \\
   & \textbf{Ours (base)}$^*$ & 0.0708 & 0.0355 & \aligntablebestresult{0.2639} & 0.0346 & 97.64 & 98.94 & 99.46  \\
   & \textbf{Ours (fusion)}$^*$ & \aligntablebestresult{0.0696} & \aligntablebestresult{0.0340} & 0.2646 & \aligntablebestresult{0.0341} & \aligntablebestresult{97.81} & \aligntablebestresult{98.98} & \aligntablebestresult{99.48}  \\
  \cmidrule[1.3pt]{2-9}
   & Bifuse~\cite{wang2020bifuse} & 0.0789 & 0.0592 & 0.2596 & 0.0443 & 95.90 & 98.38 & 99.07  \\
   & UniFuse~\cite{jiang2021unifuse} & 0.0765 & 0.0485 & 0.2802 & 0.0416 & 96.55 & 98.46 & 99.10  \\
   & \textbf{Ours (base)} & 0.0647 & 0.0347 & \tablebestresult{0.2628} & 0.0351 & 97.70 & 98.89 & 99.37  \\
   & \textbf{Ours (fusion)} & \tablebestresult{0.0631} & \tablebestresult{0.0327} & 0.2631 & \tablebestresult{0.0346} & \tablebestresult{97.82} & \tablebestresult{98.92} & \tablebestresult{99.38}  \\
  \cmidrule[1.3pt]{2-9}
   & ${\Diamond}$\textbf{Ours (base)} & 0.0572 & 0.0320 & 0.2220 & 0.0309 & 97.97 & 99.04 & 99.46  \\
   & ${\Diamond}$\textbf{Ours (fusion)} & \tablebestresult{0.0553} & \tablebestresult{0.0298} & \tablebestresult{0.2217} & \tablebestresult{0.0303} & \tablebestresult{98.09} & \tablebestresult{99.07} & \tablebestresult{99.47}  \\
  \midrule[1.5pt]
  \multirow[c]{10}{*}{3D60} & OmniDepth$^*$~\cite{zioulis2018omnidepth} & 0.1706 & 0.0931 & 0.3171 & 0.0725 & 90.92 & 97.02 & 98.51 \\
   & FCRN$^*$~\cite{laina2016fcrn} & 0.1381 & 0.0699 & 0.2833 & 0.0473 & 95.32 & 99.05 & 99.66 \\
   & \textbf{Ours (base)}$^*$ & 0.0766 & 0.0331 & 0.1802 & 0.0263 &  99.08 & \tablebestresult{99.71} & \tablebestresult{99.87}  \\
   & \textbf{Ours (fusion)}$^*$ & \tablebestresult{0.0733} & \tablebestresult{0.0316} & \tablebestresult{0.1770} & \tablebestresult{0.0257} & \tablebestresult{99.13} & \tablebestresult{99.71} & \tablebestresult{99.87}  \\
  \cmidrule[1.3pt]{2-9}
   & Bifuse~\cite{wang2020bifuse} & 0.1143 & 0.0615 & 0.2440 & 0.0428 & 96.99 & 99.27 & 99.69  \\
   & UniFuse~\cite{jiang2021unifuse} & 0.0996 & 0.0466 & 0.1968 & 0.0315 & 98.35 & 99.65 & 99.87  \\
   & OmniFusion~\cite{li2022omnifusion} & $\times$ & 0.0430 & \tablebestresult{0.1808} & 0.0735 & 98.59 & 99.69 & \tablebestresult{99.89}  \\
   & \textbf{Ours (base)} & 0.0819 & 0.0367 & 0.1823 & 0.0271 & 99.05 & \tablebestresult{99.71} & 99.87 \\
   & \textbf{Ours (fusion)} & \tablebestresult{0.0809} & \tablebestresult{0.0357} & 0.1818 & \tablebestresult{0.0270} & \tablebestresult{99.11} & \tablebestresult{99.71} & 99.87  \\
  \noalign{\smallskip}\hline
  \end{tabular}
  
\end{table*}

On Pano3D, based on results in Table~\ref{tab:3},
compared with best baseline results, our method reduces $wAbsRel$ by $\mathbf{23.0}\%$ (\textit{Fusion}) on tiny split and $\mathbf{16.3}\%$ (\textit{Fusion}) on the medium split. 
The $wRMSE$ is reduced by $\mathbf{42.1}\%$ (\textit{Fusion}) and $\mathbf{43.5}\%$ (\textit{Fusion}) on these two splits. 
We also train on GV2's medium and tested on fullplus to provide additional reference of the method's performance.
Results on Pano3D show that the proposed method has good generalization ability.
The proposed method is weaker on accuracy metrics on fullplus test splits;
this is probably because paying attention to the global context of the image makes the method fragile to large differences between training data and test data. 
We believe that more training data can solve this problem.

On synthetic datasets, the proposed method is also overall the best, as shown in Table ~\ref{tab:2}: 
comparing with previous work UniFuse~\cite{jiang2021unifuse}, 
on PSunCG datasets, the $RMSE$ decrease $\mathbf{6.5}\%$ (\textit{Fusion}), 
and the $AbsRel$ decrease $\mathbf{32.6}\%$ (\textit{Fusion}). 
However, on these two datasets, \textit{Fusion} does not outperform \textit{Base} too much. 
This is because of the local correlation between the rendered depths from 3D models and the brightness of color images~\cite{jiang2021unifuse}, 
which derives the network to use ``local'' brightness information as the main cue for predicting depth, 
thus the global context retrieving ability of \textit{Fusion} version does not contribute to final results.

\subsection{Ablation Study}
\label{sec:4.2}

\paragraph{\textbf{Different backbone networks}} 
We adapt the proposed work with different backbones and show accuracy comparison in Table~\ref{tab:4}. 
Changing the backbone of the proposed method from Swin to ResNet shows obvious performance downgrade, 
but still better than methods like ACDNet and UniFuse with less number of parameters and better latency. 
This proves that the improvement of the proposed work not only comes from the backbone we use but also the decoder on the spherical surface and our cross attention fusion module.

\begin{table}
  \centering
  \tiny
  \caption{Accuracy test results using KITTI outdoor panorama datasets~\cite{Garanderie2018Eliminating}. Evaluation method is adapted from evaluation methods of M3D datasets~\cite{chang2017matterport3d}.}
  \label{tab:4.2}   

  \begin{tabular}{c|c|c|c|c}
  \hline\noalign{\smallskip}
  \multirow[c]{2}{*}{Method} & \multicolumn{3}{c|}{Error $\Downarrow$} & Accuracy (\%) $\Uparrow$ \\
  & ${AbsRel}$ & ${RMSE}$ & ${RMSLE}$ & $\delta < 1.25$ \\
  \noalign{\smallskip}\hline
  \hline\noalign{\smallskip}
  OmniFusion~\cite{li2022omnifusion} & 0.0977 & 4.536 & 0.193 & 90.80 \\
  UniFuse~\cite{jiang2021unifuse} & 0.0945 & 4.186 & 0.076 & 92.58 \\
  Proposed & \tablebestresult{0.0657} & \tablebestresult{3.822} & \tablebestresult{0.060} & \tablebestresult{93.60} \\
  \noalign{\smallskip}\hline
  \end{tabular}
  
\end{table}

\paragraph{\textbf{Outdoor datasets}} 
We evaluate the method's performance on KITTI outdoor datasets~\cite{Garanderie2018Eliminating} using our own train/test split, 
and results are shown in Table~\ref{tab:4.2}.
Although the proposed method aims for solving depth estimation mostly for indoor scenes, 
its capability of capturing global context helps the proposed method get overall best results.

\begin{table}
  \centering
  \tiny
  \caption{Effects of error angles when the panorama images are not optimally aligned to gravity. Data with different tilt angles is retrieved by re-rendering from depth images from M3D datasets~\cite{chang2017matterport3d}.  }
  \label{tab:7}   

  \begin{tabular}{c|c|c|c|c|c}
  \hline\noalign{\smallskip}
  \multirow[c]{2}{*}{Methods} & \multirow[c]{2}{*}{\tabincell{c}{Tilt \\ angle}} & \multicolumn{3}{c|}{Error $\Downarrow$} & Accuracy (\%) $\Uparrow$ \\
  & & ${AbsRel}$ & ${RMSE}$ & ${RMSLE}$ & $\delta < 1.25$\\
  \noalign{\smallskip}\hline
  \hline\noalign{\smallskip}
  \multirow[c]{3}{*}{Bifuse~\cite{wang2020bifuse}} & ${0}^{\circ}$ & 0.2048 & 0.6259 & 0.1134 & 84.52 \\
  & ${2}^{\circ}$ & 0.3888 & 0.9805 & 0.1852 & 61.44 \\
  & ${5}^{\circ}$ & 0.4905 & 1.0225 & 0.2250 & 54.40 \\
   \noalign{\smallskip}\hline\noalign{\smallskip}
  \multirow[c]{3}{*}{SliceNet~\cite{pintore2021slicenet}} & ${0}^{\circ}$ & 0.1764 & 0.6133 & 0.1045 & 87.16 \\
  & ${2}^{\circ}$ & 0.2645 & 0.7026 & 0.1334 & 72.56 \\
  & ${5}^{\circ}$ & 0.3032 & 0.7720 & 0.1482 & 68.79 \\
   \noalign{\smallskip}\hline\noalign{\smallskip}
  \multirow[c]{3}{*}{\textbf{Ours}} & ${0}^{\circ}$ & 0.0865 & 0.4052 & 0.0559 & 92.64 \\
  & ${2}^{\circ}$ & 0.0927 & 0.4282 & 0.0600 & 91.60 \\
  & ${5}^{\circ}$ & 0.1091 & 0.4792 & 0.0711 & 88.14 \\
  \noalign{\smallskip}\hline
  \end{tabular}
  
\end{table}

\paragraph{\textbf{Imperfect alignment to gravity direction}} This may happen when the camera is not placed on a horizontal plane. 
Results in Table~\ref{tab:7} demonstrate the performance downgrade of the proposed method under different tilt angles between the axis of the panorama image and gravity direction; 
ground truth used for evaluation is retrieved by re-rendering from depth images from the M3D dataset. For tilt angles $2^{\circ}$ and $5^{\circ}$, the ${RMSE}$ of the proposed method increase $\mathbf{5.6}\%$ and $\mathbf{18.3}\%$ respectively, while performance downgrades for SliceNet~\cite{pintore2021slicenet} are $\mathbf{14.6}\%$ and $\mathbf{25.9}\%$ respectively.
This proves that the proposed method process better capability of resisting the jeopardization from imperfect gravity alignment.

\section{Conclusions}
\label{sec:5}

This paper proposes a novel method for panorama depth estimation. 
By adapting uniform sampling on the unit sphere, 
and by using self-attention calculation to enhance network's capability of retrieving global context, 
the method achieved impressive accuracy on all publicly available datasets. 
However, the proposed method can still be further improved: 
by moving the encoder parts onto unit sphere, the method will possess better distortion-removal capability, which can improve the accuracy of the proposed method;
edge-preserving filters can be utilized in interpolation process of $\mathbf{f}_{{\Pi}\rightarrow{\mathbf{S^{2}}}}$ and $\mathbf{f}_{{\mathbf{S^{2}}}\rightarrow\Pi}$ to further improve depth estimation quality on edges.

\bibliographystyle{IEEEtran}
\bibliography{camera_ready}
\end{document}